\documentclass[runningheads]{llncs}

\usepackage{multirow}

\usepackage{eccvabbrv}

\usepackage{graphicx}
\usepackage{booktabs}

\usepackage[accsupp]{axessibility}  

\usepackage{hyperref}

\usepackage{orcidlink}

\begin{document}

\title{DiffInf: Influence-Guided Diffusion for Supervision Alignment in
Facial Attribute Learning} 

\titlerunning{DiffInf}

\author{Basudha Pal\inst{1}\orcidlink{0009-0009-0920-8565} \and
Zhaoyang Wang\inst{2}\orcidlink{0009-0008-7003-1909} \and
Rama Chellappa\inst{1,3}\orcidlink{0000-0002-7638-1650}}

\authorrunning{B.~Pal et al.}

\institute{Department of Electrical and Computer Engineering, Johns Hopkins University, Baltimore, MD, USA\and
Department of Computer Science, Johns Hopkins University, Baltimore, MD, USA
\and
Department of Biomedical Engineering, Johns Hopkins University, Baltimore, MD, USA\\
\email{\{bpal5, zwang303, rchella4\}@jhu.edu}}

\maketitle

\begin{abstract}

  Facial attribute classification relies on large-scale annotated datasets in which traits such as age and expression are inherently ambiguous and continuous but discretized into categorical labels. Annotation inconsistencies arise from subjectivity and visual confounders including pose, illumination, expression, and demographic variation, creating mismatches between images and assigned labels. These errors impair representation learning and degrade downstream prediction. We introduce DiffInf, a self-influence--guided diffusion framework for mitigating annotation inconsistencies in facial attribute learning. We first train a baseline classifier and compute sample-wise self-influence scores using a practical first-order approximation to identify training instances that disproportionately destabilize optimization. Rather than discarding these influential samples, we apply targeted generative correction with a latent diffusion autoencoder to better align visual content with assigned labels while preserving identity and realism. To enable differentiable guidance, we train a lightweight predictor of high-influence membership and use it as a surrogate influence regularizer. The edited samples replace the originals, yielding an influence-refined dataset of unchanged size. Across multi-class facial attribute classification, DiffInf consistently improves generalization over standard noisy-label training, robust optimization baselines, and influence-based filtering. Our results show that repairing influential annotation inconsistencies at the image level improves downstream facial attribute classification without sacrificing distributional coverage.

\end{abstract}

\section{Introduction}
\label{sec:intro}

Deep neural networks underpin modern face analysis systems for facial attribute prediction, demographic inference, and biometric understanding across security, healthcare, and human--computer interaction. Their performance, however, is fundamentally constrained by the semantic consistency of supervisory signals. Many facial attributes, including age and other soft biometric traits, are inherently ambiguous and continuous yet discretized into multi-class categories. Their visual manifestation is influenced by confounders such as illumination, ethnicity, cosmetics, lifestyle, and pose, leading to annotation uncertainty and label noise in large-scale datasets \cite{Northcutt2021}, \cite{Xiao2015}, \cite{Patrini2017}, \cite{Jiang2018}. When image appearance and assigned labels diverge, models receive contradictory supervision, degrading representation learning, generalization, calibration, and fairness \cite{Song2022}, \cite{Buolamwini2018}, \cite{Kortylewski2019Bias}.

Data attribution and influence estimation provide tools to quantify how individual training samples shape model behavior. Classical influence functions \cite{Koh2017} and scalable approximations \cite{Pruthi2020}, \cite{park2023trak} estimate each training instance's contribution to optimization dynamics and predictions. These approaches have been used to analyze memorization \cite{Feldman2020}, detect mislabeled or bias-inducing samples \cite{Northcutt2021}, enable selective forgetting \cite{Golatkar2020}, and improve robustness under noisy supervision \cite{Han2018}, \cite{Chen2019}, \cite{Zhang2021DivideMix}. A consistent finding is that a small subset of highly influential samples can disproportionately govern the learned decision surface, particularly under annotation ambiguity \cite{Feldman2020}, \cite{Schwinn2023InfluenceSurvey}. Existing methods typically respond through removal, reweighting, or relabeling \cite{Ilyas2022Datamodels}, \cite{Ghorbani2019}, leaving open whether influential image--label inconsistencies can instead be addressed through direct generative refinement.

Removal-based strategies can deplete the training distribution. Discarding a high-influence sample due to supervision mismatch removes not only a corrupted label but also a visually valid instance that may encode rare covariate combinations important for representing the full data manifold. By shifting from exclusion to semantic conservation, we treat such samples as informational assets requiring alignment rather than supervisory noise to be eliminated. This perspective preserves diversity and rare modes while stabilizing optimization through more consistent supervision.

In parallel, diffusion models have emerged as state-of-the-art generative frameworks \cite{Ho2020}, \cite{Rombach2022}, \cite{Nichol2021}, demonstrating strong realism and controllability, including applications in bias mitigation \cite{pal2024diversinet}, \cite{pal2024gamma}, privacy preservation \cite{liu2023diffprotect}, face swapping \cite{kim2025diffface}, and related facial tasks \cite{boutros2023idiff}, \cite{huang2023collaborative}, \cite{ponglertnapakorn2023difareli}. Beyond unconditional synthesis, diffusion enables identity-preserving semantic editing via partial denoising \cite{Meng2022}, cross-attention control \cite{Hertz2022}, CLIP-guided manipulation \cite{Kim2022}, and instruction- or mask-based editing \cite{Brooks2023InstructPix2Pix}, \cite{Couairon2023DiffEdit}. These advances suggest that generative models can perform localized semantic repair by aligning visual appearance with assigned attribute labels.

A naive refinement strategy would use a pretrained facial-attribute classifier to guide diffusion. However, this creates a circular dependency. If the guidance model is trained on the same task as the downstream model, it inherits the same biases and errors \cite{Buolamwini2018}, \cite{Zhao2017Bias} that the refinement process is intended to correct, potentially reinforcing supervisory mistakes. Moreover, a static classifier does not identify which images most destabilize training for the learner of interest. This motivates a model-aware framework in which influence determines \emph{which} samples to correct, and diffusion determines \emph{how} to correct them.

We introduce DiffInf, a self-influence--guided diffusion framework for facial attribute learning under noisy supervision. DiffInf is attribute-agnostic and applicable to multi-class attributes with ambiguous visual boundaries. We evaluate it on multi-class facial age prediction and facial expression recognition, building upon established formulations in the field \cite{Rothe2015}, \cite{Gong2019}, \cite{Levi2015}, \cite{Niu2016}. A baseline classifier is first trained, after which self-influence scores are computed using a first-order checkpoint-based approximation \cite{Pruthi2020} to quantify each sample's impact on training dynamics. High-influence samples with potential image--label inconsistencies are then targeted for correction. Rather than removing them, DiffInf performs generative supervision refinement using a pretrained latent diffusion autoencoder \cite{preechakul2022diffusion}. We optimize latent variables to produce identity-preserving, visually realistic samples that are less disruptive to optimization under their assigned labels. The edited images replace the original influential samples, producing an influence-refined training distribution of unchanged size. Across tasks, influence-guided generative replacement consistently improves generalization, robustness, and calibration. DiffInf complements recent influence-aware robust learning approaches \cite{pal2025grasp} using influence as a localization signal for direct data-level correction rather than suppressing the effects of noisy samples during optimization. The main contributions of this study are summarized as follows:

\begin{itemize}
\item We introduce DiffInf, a self-influence--guided diffusion framework that aligns training images with assigned facial attribute labels under noisy supervision.
\item We propose targeted generative replacement of high-influence samples via latent diffusion autoencoding, producing an influence-refined dataset while preserving data coverage.
\item We incorporate a differentiable predictor of high-influence membership to enable scalable influence-guided correction during latent optimization.
\item We demonstrate consistent improvements in facial attribute prediction, particularly for multi-class age and expression classification, and show that repairing influential mismatches is more effective than simply removing them.
\end{itemize}

\begin{figure}
  \centering
    \includegraphics[width=\linewidth, keepaspectratio]{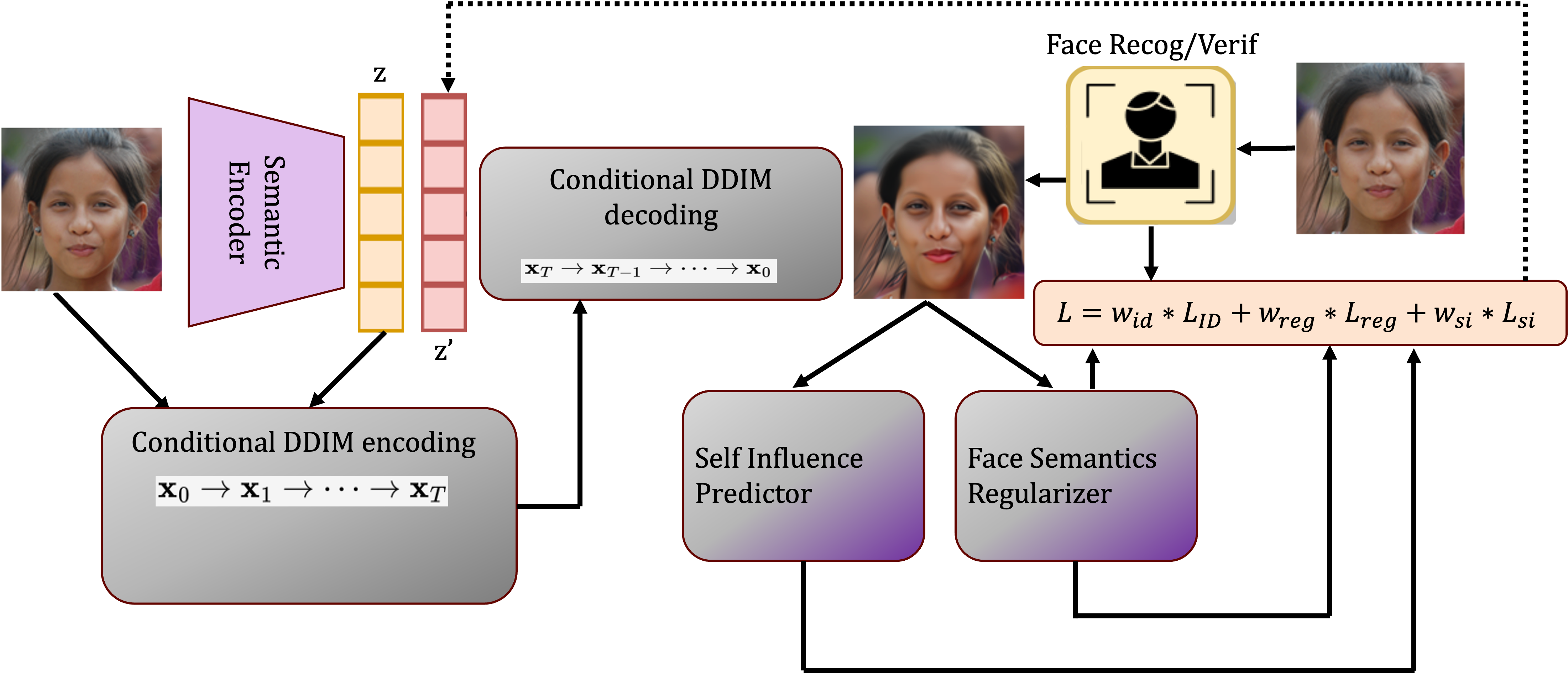}
    \caption{The DiffInf framework for influence-guided generative supervision alignment. The process begins with a semantic encoder and conditional DDIM encoding to project the input image into a latent representation. During decoding, iterative latent optimization is guided by a composite loss balancing identity preservation (via a face recognition/verification network), influence suppression (via a predictor of high-influence membership), and semantic alignment while maintaining realism. By targeting high-influence samples for generative correction, DiffInf realigns training images with their assigned attribute labels while preserving identity fidelity and structural realism.}
    \label{fig:pipeline}
\end{figure}

\section{Proposed Method}
\label{sec:method}

Our approach, DiffInf (Fig.~\ref{fig:pipeline}), addresses facial attribute learning under noisy supervision by explicitly repairing high-impact training samples through influence-guided generative correction. Instead of treating all samples uniformly or relying on heuristic filtering strategies, DiffInf first quantifies the self-influence of each training instance using a scalable first-order approximation of classical influence functions. This produces a model-aware estimate of how strongly each sample shapes the learned decision boundary.

Samples exhibiting anomalously high self-influence often arise from annotation inconsistencies, semantic ambiguity, or image--label mismatch. Rather than discarding such instances, DiffInf performs targeted generative correction. Specifically, high-influence samples are reconstructed through identity-preserving diffusion editing that encourages alignment between visual appearance and assigned labels while maintaining intrinsic semantic structure. The corrected images replace the original samples, yielding an influence-refined dataset. Retraining the classifier on this refined dataset leads to improved generalization and more stable optimization. Unlike standard augmentation, outlier pruning, or classifier-guided editing alone, DiffInf tightly couples causal data attribution with generative supervision alignment. As a result, correction is applied precisely where it most affects learning dynamics.

\subsection{Problem Setup}

We consider supervised facial attribute learning under noisy labels. Let $
\mathcal{D} = \{(x_i,\tilde{y}_i)\}_{i=1}^{N}
$
denote a training dataset, where $x_i \in \mathbb{R}^{H \times W \times 3}$ is a facial image and $\tilde{y}_i \in \{1,\dots,C\}$ is a potentially corrupted categorical attribute label. The unknown clean label is denoted by $y_i$.
A facial attribute classifier
$
f_\theta : \mathbb{R}^{H \times W \times 3} \rightarrow \Delta^{C}
$
with parameters $\theta$ is trained by minimizing the empirical risk
\begin{equation}
\mathcal{L}_{\text{train}}(\theta)
=
\frac{1}{N}\sum_{i=1}^{N}
\ell\!\left(f_\theta(x_i), \tilde{y}_i\right),
\end{equation}
where $\ell(\cdot,\cdot)$ denotes cross-entropy loss.

This formulation implicitly assumes label reliability. However, when $\tilde{y}_i$ contradicts the semantic content of $x_i$, gradient updates introduce inconsistent optimization signals. Such samples frequently retain large gradients even near convergence, distorting the learned decision boundary and degrading generalization. DiffInf addresses this issue by identifying and repairing these destabilizing samples rather than suppressing their gradients or removing them from the dataset.

\subsection{Influence from First Principles}

Training samples contribute unequally to optimization. Influence functions quantify how infinitesimal perturbations to a training point affect the learned parameters \cite{Koh2017}. Let
$
\hat{\theta}
=
\arg\min_\theta
\frac{1}{N} \sum_{j=1}^{N} \ell_j(\theta),
\quad
\ell_j(\theta)
=
\ell(f_\theta(x_j),\tilde{y}_j).
$
Upweighting a single sample $z_i=(x_i,\tilde{y}_i)$ by $\epsilon$ yields:

\begin{equation}
\hat{\theta}^{(i)}_\epsilon
=
\arg\min_\theta
\left(
\frac{1}{N} \sum_{j=1}^{N} \ell_j(\theta)
+
\epsilon \ell_i(\theta)
\right).
\end{equation}

Differentiating at $\epsilon=0$ gives

\begin{equation}
\frac{d\hat{\theta}^{(i)}_\epsilon}{d\epsilon}\Big|_{\epsilon=0}
=
- H_{\hat{\theta}}^{-1}
\nabla_\theta \ell_i(\hat{\theta}),
\end{equation}

where

\[
H_{\hat{\theta}}
=
\frac{1}{N}
\sum_{j=1}^{N}
\nabla_\theta^2 \ell_j(\hat{\theta})
\]

is the empirical Hessian. This expression reveals that parameter sensitivity depends both on the gradient magnitude of the sample and on the curvature of the loss landscape.
DiffInf focuses on self-influence:

\begin{equation}
\mathcal{I}_{\text{self}}(z_i)
=
\nabla_\theta \ell_i(\hat{\theta})^\top
H_{\hat{\theta}}^{-1}
\nabla_\theta \ell_i(\hat{\theta}).
\end{equation}

Self-influence measures how strongly upweighting a sample perturbs the learned parameters and thus its own contribution to the training objective. Near convergence, well-aligned samples typically produce small gradients and low self-influence, whereas mislabeled or semantically inconsistent samples retain large gradients and therefore exhibit high self-influence.

\subsection{Practical Self-Influence Approximation}

Computing $H_{\hat{\theta}}^{-1}$ is infeasible for modern deep networks. We therefore adopt a checkpoint-based first-order approximation inspired by TracIn:

\begin{equation}
\tilde{\mathcal{I}}_{\text{self}}(z_i)
=
\sum_{t \in \mathcal{T}}
\eta_t
\left\langle
\nabla_\theta \ell_i(\theta^{(t)}),
\nabla_\theta \ell_i(\theta^{(t)})
\right\rangle.
\end{equation}

Here $\theta^{(t)}$ denotes intermediate checkpoints during training and $\eta_t$ the learning rate at step $t$. This accumulated gradient energy reflects how persistently a sample induces large parameter updates throughout optimization. In practice, we compute this score across saved checkpoints to obtain a stable proxy for self-influence. The high-influence subset is defined as
$
\mathcal{H}
=
\left\{
i
\mid
\tilde{\mathcal{I}}_{\text{self}}(z_i)
\text{ is in the top } \tau\%
\right\}.
$
Only samples in $\mathcal{H}$ are considered for generative correction.

\subsection{Learning an Influence Predictor}

Directly computing influence during generative optimization would require repeated gradient evaluations, which is computationally expensive. To enable scalable correction, we train a lightweight influence prediction network $h_\omega$.
Let $g_i$ denote features extracted from the backbone $f_\theta$. The predictor learns $h_\omega(g_i, \tilde{y}_i)
\rightarrow p_{\text{high}}(z_i)$, the probability that sample $z_i$ belongs to the high-influence subset $\mathcal{H}$. Supervision is obtained from thresholded influence scores, and the network is trained using binary cross-entropy. This model acts as a differentiable surrogate for high-influence membership. Importantly, it does not infer the true semantic label; instead, it approximates which samples destabilize training under their assigned supervision.

\subsection{Diffusion Autoencoders}

To repair high-influence samples, DiffInf employs a diffusion autoencoder composed of a semantic encoder $E_\phi$ and a conditional diffusion decoder $D_\psi$. Given an image $x_0$, $s = E_\phi(x_0)$, where $s$ encodes identity-consistent semantic information. The diffusion model is conditioned on the assigned label $c=\tilde{y}_i$. The forward diffusion process is

\begin{equation}
x_t
=
\sqrt{\bar{\alpha}_t}\,x_0
+
\sqrt{1-\bar{\alpha}_t}\,\epsilon,
\quad
\epsilon \sim \mathcal{N}(0,I).
\end{equation}

We first estimate the denoised image

\begin{equation}
\hat{x}_0 =
\frac{x_t - \sqrt{1-\bar{\alpha}_t}\,\epsilon_\psi(x_t,s,c)}
{\sqrt{\bar{\alpha}_t}},
\end{equation}

and then compute the DDIM reverse update

\begin{equation}
\begin{aligned}
x_{t-1}
&=
\sqrt{\bar{\alpha}_{t-1}}\,\hat{x}_0
+
\sqrt{1-\bar{\alpha}_{t-1}}\,
\epsilon_\psi(x_t,s,c).
\end{aligned}
\end{equation}

Optimization is performed in latent space, ensuring edits remain constrained by the learned generative prior.

\subsection{DiffInf: Influence-Guided Generative Correction}

For each high-influence sample $x_i \in \mathcal{H}$, DiffInf generates a corrected image $\hat{x}_i$ that remains identity-consistent with the original image while reducing its destabilizing impact on training. Intuitively, the corrected sample should resemble the same individual but exhibit visual attributes that are more consistent with the assigned label $\tilde{y}_i$. To achieve this, generative reconstruction is guided by a composite objective that balances identity preservation, perceptual consistency, and influence suppression. The optimization objective is

\begin{equation}
\hat{x}_i
=
\arg\min_{x}\;
w_{\mathrm{id}}\mathcal{L}_{\mathrm{id}}(x, x_i)
+
w_{\mathrm{reg}}\mathcal{L}_{\mathrm{reg}}(x, x_i)
+
w_{\mathrm{si}}\mathcal{L}_{\mathrm{si}}(x,\tilde{y}_i).
\end{equation}

The three terms in this objective enforce complementary constraints on the generated image.

\paragraph{Identity Preservation}

\begin{equation}
\mathcal{L}_{\mathrm{id}}(x, x_i)
=
1 -
\cos\!\big(F_{\mathrm{id}}(x), F_{\mathrm{id}}(x_i)\big).
\end{equation}

This term ensures that the corrected image retains the identity of the original subject. Rather than enforcing strict pixel-level similarity, identity consistency is measured in a learned facial embedding space using a pretrained recognition network $F_{\mathrm{id}}$. By minimizing cosine distance between embeddings, the optimization preserves facial characteristics such as bone structure, facial geometry, and identity-specific features while allowing attribute-relevant appearance changes.
\paragraph{Regularization}

\begin{equation}
\mathcal{L}_{\mathrm{reg}}(x, x_i)
=
\lambda_{\mathrm{struct}}\mathcal{L}_{\mathrm{struct}}(x, x_i)
+
\lambda_{\mathrm{perc}}\mathcal{L}_{\mathrm{perc}}(x, x_i).
\end{equation}

Regularization terms stabilize the generative optimization and prevent unrealistic edits. The structural component $\mathcal{L}_{\mathrm{struct}}$ enforces consistency of facial layout using a pretrained face parsing network, ensuring that key regions such as eyes, nose, and mouth maintain coherent spatial structure. The perceptual component $\mathcal{L}_{\mathrm{perc}}$ (e.g., LPIPS and reconstruction losses) encourages visual similarity to the original image in feature space. Together, these terms constrain the edit to remain within a realistic perceptual neighborhood of the input while permitting attribute-relevant adjustments.

\paragraph{Self-Influence Suppression}

\begin{equation}
\mathcal{L}_{\mathrm{si}}(x,\tilde{y}_i)
=
h_\omega(g(x), \tilde{y}_i).
\end{equation}

The influence term directly targets the root cause of training instability. The predictor $h_\omega$ estimates the probability that a sample belongs to the high-influence subset $\mathcal{H}$. Minimizing this term encourages the corrected sample to move toward regions of feature space associated with low self-influence, making it easier for the classifier to integrate into the learned representation. Consequently, the corrected image behaves like a typical, label-consistent training instance while retaining the identity and visual characteristics of the original image.

\subsection{Refined Training Distribution}

After correction,

\begin{equation}
\mathcal{D}'
=
\Big(
\mathcal{D}
\setminus
\{(x_i,\tilde{y}_i): i \in \mathcal{H}\}
\Big)
\cup
\{(\hat{x}_i,\tilde{y}_i): i \in \mathcal{H}\}.
\end{equation}

This procedure preserves dataset cardinality and coverage while removing destabilizing supervision inconsistencies. DiffInf therefore transforms influence estimation from a passive analytical tool into an active generative intervention that improves both optimization stability and downstream generalization.

\subsection{Experimental Setup}

For age classification, we construct three semantically separated categories: Young, Middle, and Old. For expression recognition, we consider four discrete facial expressions: happy, neutral, surprised, and sad. All images are resized to $128 \times 128$ and normalized to match the pretrained diffusion autoencoder.
To simulate annotation corruption, symmetric label noise is introduced into the training set. A fixed proportion $n\%$ of training samples is randomly selected, and each selected label is replaced with a uniformly sampled incorrect label from the remaining classes. Validation and test sets remain clean throughout. Unless otherwise specified, experiments use 30\% label noise for age classification and 20\% for expression recognition. These settings were selected based on influence-estimation stability. Since age groups discretize a continuous attribute into coarse bins, 30\% noise preserves sufficient semantic structure while introducing enough inconsistencies for meaningful influence localization. In contrast, expression recognition exhibits more rapid degradation of class boundaries under heavier corruption, reducing the reliability of checkpoint-based influence estimation.

\begin{figure}[t]
    \centering
    \includegraphics[width=\linewidth]{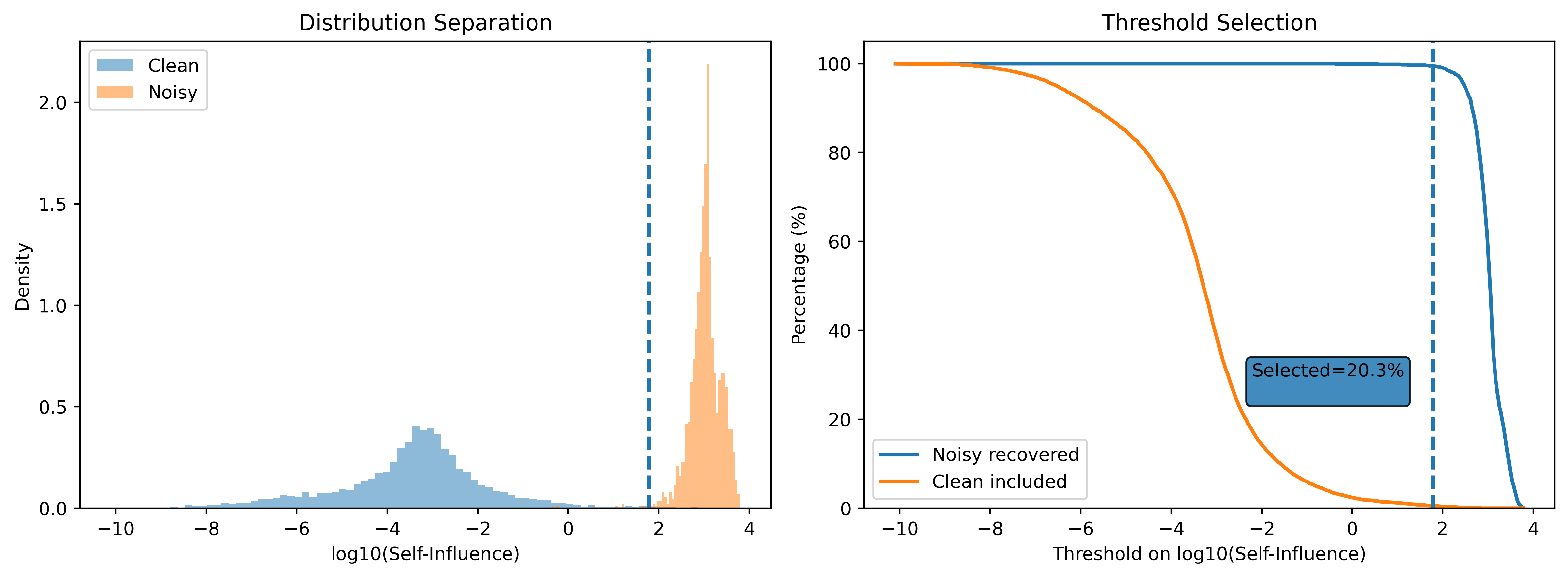}
    \caption{
   Empirical validation of self-influence under synthetic label noise. \textbf{Left:} Distribution of log self-influence values for clean and corrupted training samples after classifier training. Corrupted samples consistently exhibit higher self-influence, indicating persistent optimization instability caused by annotation errors. \textbf{Right:} Recovery of corrupted samples and inclusion of correctly labeled samples as the self-influence threshold varies. Selecting approximately the top 20\% highest self-influence samples recovers nearly all corrupted samples while retaining only a small fraction of correctly labeled instances.
    }
    \label{fig:threshold}
\end{figure}

A ResNet-18 classifier with a two-layer fully connected head is first trained on the noisy training set using cross-entropy loss and Adam with a learning rate of $1\times10^{-4}$ for 100 epochs. Model checkpoints are saved throughout training to compute the first-order self-influence approximation described in Sec.~\ref{sec:method}. Before performing generative correction, we verify that self-influence provides a reliable localization signal for optimization-disruptive samples. Since the corrupted indices are known under synthetic label noise, we can directly compare the self-influence distributions of clean and corrupted samples after training the baseline classifier.
Figure~\ref{fig:threshold}(left) shows a clear separation between the two groups. Correctly labeled samples concentrate at low self-influence values, whereas corrupted samples consistently exhibit substantially higher influence. This behavior is expected, as visually inconsistent labels remain difficult for the classifier to optimize, producing persistently large gradients and consequently high self-influence.

In practice, neither the corruption rate nor the identities of noisy samples are known. Consequently, sample selection cannot rely on prior knowledge of the true noise level. Figure~\ref{fig:threshold}(right) therefore evaluates the recovery characteristics of self-influence as the selection threshold varies for age classification. As progressively larger fractions of the highest-influence samples are selected, recovery of corrupted samples rapidly saturates while the inclusion of correctly labeled samples increases much more gradually. 
Based on this trade-off, we select approximately the top 20\% highest self-influence samples for age classification, which recovers nearly all optimization-disruptive samples while retaining relatively few correctly labeled instances. The same influence-thresholding procedure is independently applied to expression recognition, with the operating point determined from its corresponding self-influence distribution. For brevity, only the age analysis is shown.

The selected samples are designated as candidates for generative correction. A binary classifier is trained on influence-derived pseudo-labels from a held-out partition to predict high-influence membership and is subsequently frozen to provide a differentiable guidance signal during diffusion-based optimization. For each selected sample, latent optimization is performed using a pretrained diffusion autoencoder \cite{preechakul2022diffusion}. The generator parameters remain fixed, with only the latent variables optimized, preserving both the learned data manifold and the dataset cardinality. After correction, the attribute classifier is retrained from scratch on the modified training set. We compare our approach against (i) a baseline trained directly on the noisy data and (ii) an influence-based filtering strategy that removes high self-influence samples before retraining.

\begin{figure}[t]
  \centering
    \includegraphics[width=\linewidth, keepaspectratio]{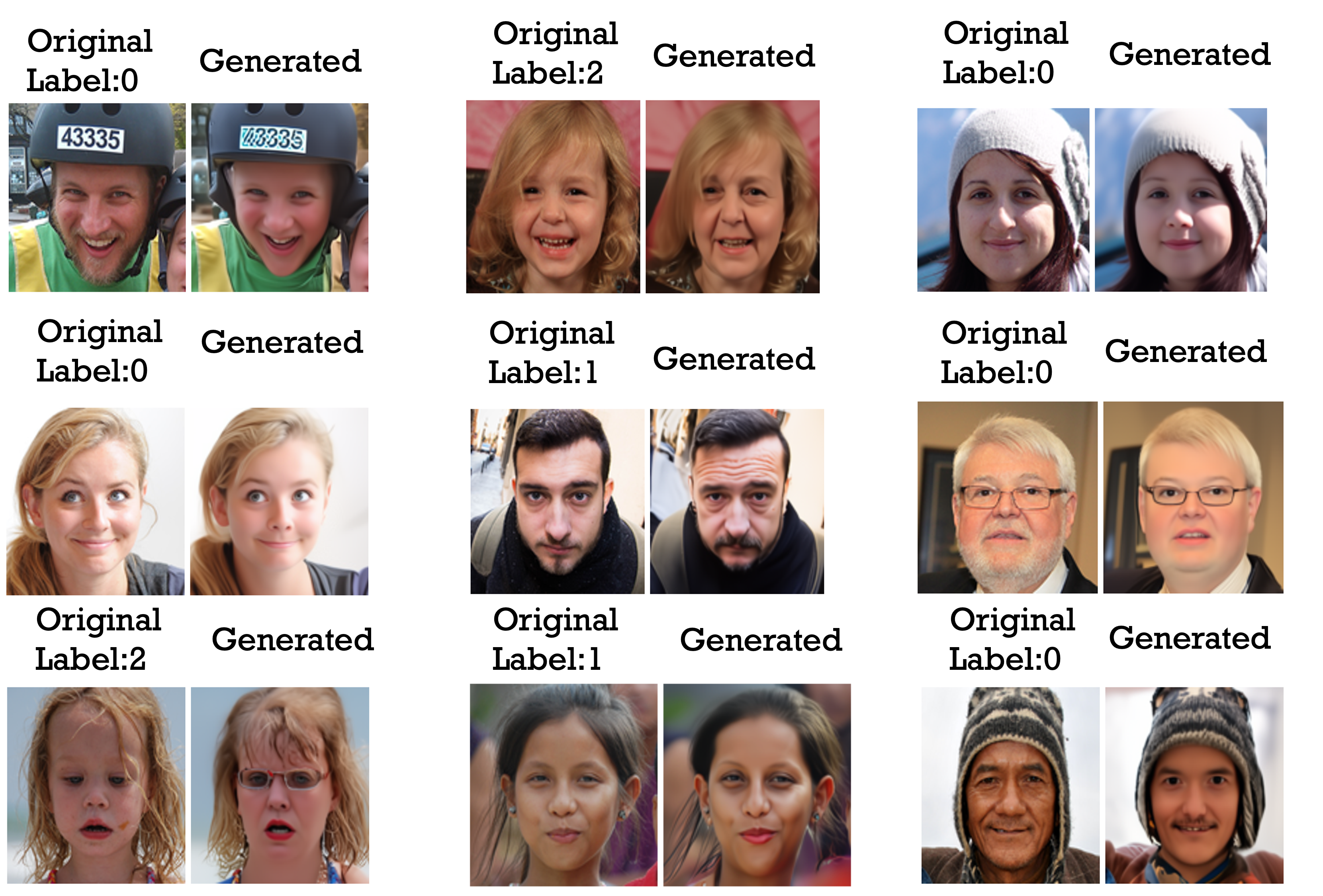}
    \caption{Qualitative examples of influence-guided generative correction (DiffInf) under noisy supervision. Each pair shows a high self-influence training sample (left) and its generated correction (right). Labels correspond to the dataset-provided age-group class, where 0 = Young (0--18), 1 = Middle (25--40), and 2 = Old (50+). Several original samples exhibit semantic mismatch with their assigned labels (e.g., an older-looking face labeled as Young). DiffInf preserves facial identity, pose, and illumination while selectively modifying age-related features to align with the assigned class, producing label-consistent samples while maintaining realism and identity fidelity.}
    \label{fig1}
    \vspace{-3mm}
\end{figure}
\begin{figure}[t]
  \centering
    \includegraphics[width=\linewidth, keepaspectratio]{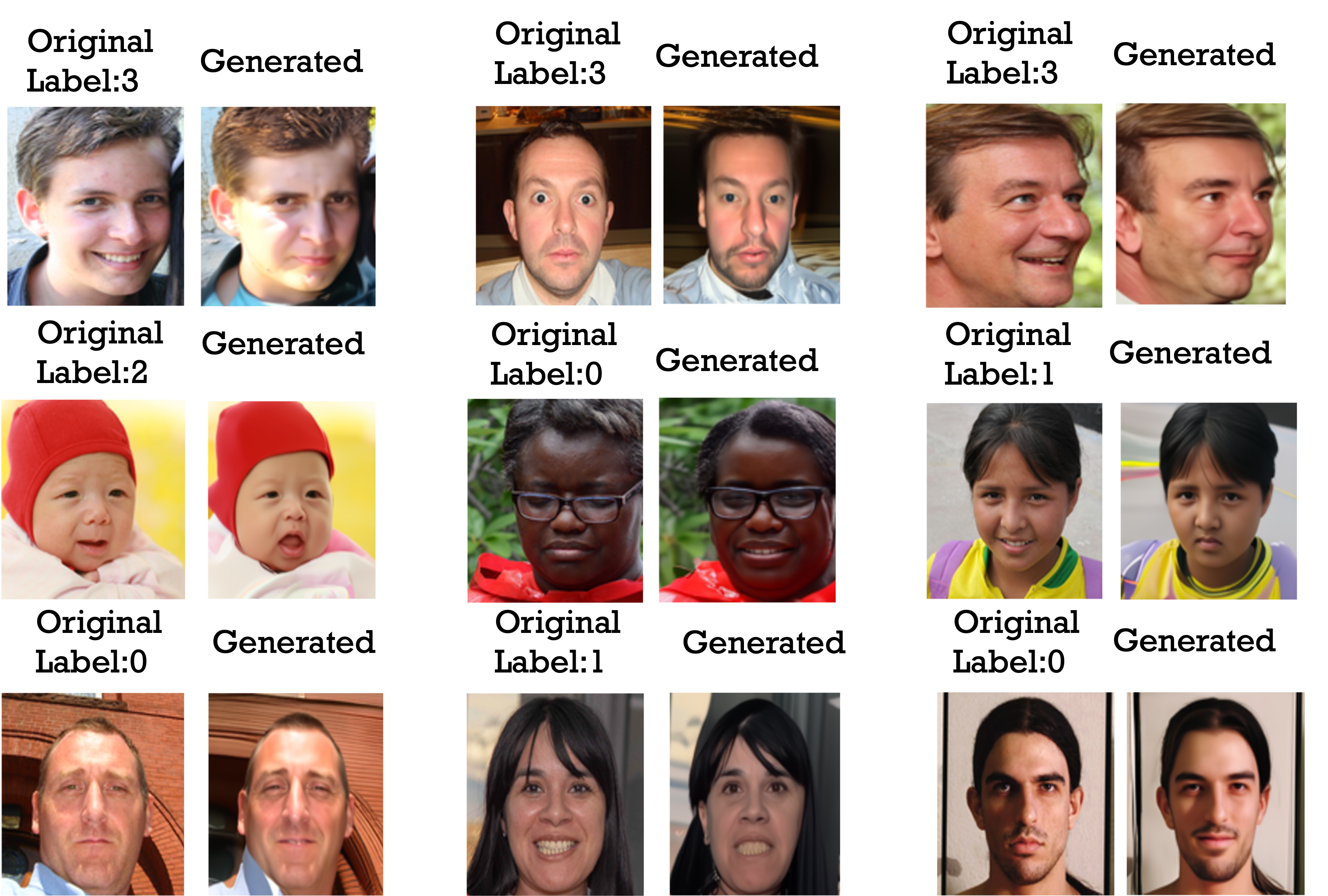}
   \caption{Qualitative examples of influence-guided generative correction (DiffInf) for facial expression classification under noisy supervision. Each pair shows an original high self-influence training sample (left) and its generated correction (right). Labels correspond to the dataset-provided expression class, where 0 = Happy, 1 = Neutral, 2 = Surprised, and 3 = Sad. Several original samples are semantically inconsistent with their assigned labels, resulting in high influence during training. DiffInf preserves facial identity, pose, and illumination while selectively modifying expression-related features (e.g., mouth, eyebrows, and eyes) to align with the assigned class, producing label-consistent samples without altering identity.}
    \label{fig2}
    \vspace{-3mm}
\end{figure}
\begin{table*}[t]
\centering
\caption{Performance on FFHQ under 30\% symmetric label noise for age classification and 20\% for expression classification. Metrics are evaluated on the clean test set. Best results are shown in bold.}
\label{tab:ffhq_classification}
\setlength{\tabcolsep}{4pt}
\begin{tabular}{llccc|ccc}
\toprule
& &
\multicolumn{3}{c|}{\textbf{Age (3-class)}} &
\multicolumn{3}{c}{\textbf{Expression (4-class)}} \\
\cmidrule(lr){3-5}
\cmidrule(lr){6-8}
\textbf{Dataset} &
\textbf{Method} &
Acc &
AUROC &
$\kappa$ &
Acc &
AUROC &
$\kappa$ \\
\midrule

\multirow{7}{*}{FFHQ}
& Original noisy        & 70.44 & 85.48 & 0.58 & 78.95 & 94.43 & 0.69 \\
& Small-loss            & 74.92 & 88.54 & 0.68 & 83.55 & 96.84 & 0.77 \\
& ELR+                  & 80.56 & 92.40 & 0.76 & 81.90 & 93.98 & 0.73 \\
& Self-inf removal      & 81.23 & 93.73 & 0.76 & 93.42 & 99.27 & 0.88 \\
& \textbf{Self-inf gen (Ours)}
                          & \textbf{83.37} & \textbf{94.94} & \textbf{0.78}
                          & \textbf{94.24} & \textbf{99.38} & \textbf{0.90} \\
& ProSelf               & 72.50 & 87.92 & 0.60 & 81.57 & 96.27 & 0.77 \\
& ProMix                & 76.55 & 90.18 & 0.68 & 78.29 & 94.56 & 0.70 \\
\bottomrule
\end{tabular}
\end{table*}
\begin{figure}[t]
    \centering
    \includegraphics[width=0.7\linewidth]{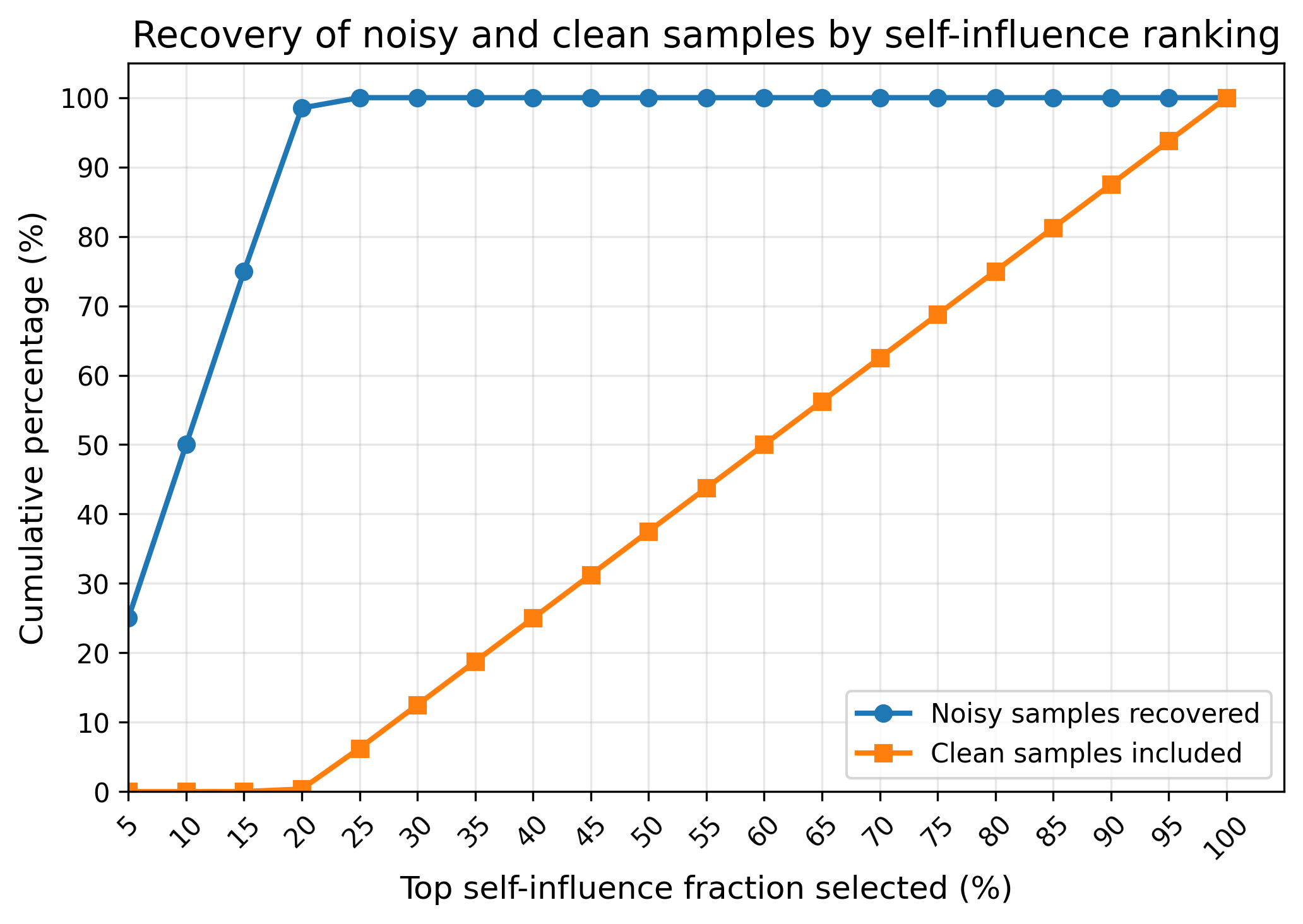}
    \caption{
    Recovery of corrupted samples as progressively larger fractions of the highest self-influence training samples are selected. The blue curve shows the cumulative percentage of corrupted samples recovered, while the orange curve denotes the cumulative proportion of correctly labeled samples included. Nearly all corrupted samples are concentrated within the top 20\% of the self-influence ranking, indicating that optimization-disruptive samples occupy the highest influence regions and motivating targeted generative correction.
    }
    \label{fig:ranking}
\end{figure}
\begin{table}[t]
\centering
\caption{Class-wise LPIPS values between original and DiffInf-generated samples. Lower values indicate higher perceptual similarity.}
\label{tab:ffhq_lpips}
\begin{tabular}{lc}
\toprule
\textbf{Class} & \textbf{LPIPS} \\
\midrule
\multicolumn{2}{l}{\textbf{Age}} \\
Young (0--18)   & 0.227 \\
Middle (25--40) & 0.219 \\
Old (50+)       & 0.224 \\
\addlinespace
\multicolumn{2}{l}{\textbf{Expression}} \\
Happy           & 0.244 \\
Sad             & 0.229 \\
Neutral         & 0.196 \\
Surprised       & 0.222 \\
\bottomrule
\end{tabular}
\vspace{-5mm}
\end{table}
\section{Results and Discussion}

Table~\ref{tab:ffhq_classification} reports quantitative performance on FFHQ under synthetic label noise for both age (3-class) and expression (4-class) recognition. Training directly on noisy labels yields 70.44\% accuracy, 85.48 AUROC, and $\kappa = 0.58$ for age, and 78.95\% accuracy, 94.43 AUROC, and $\kappa = 0.69$ for expression. These results confirm that supervision mismatch substantially degrades learning, particularly for age prediction, where discretizing a continuous attribute into coarse categories increases sensitivity to contradictory labels.

To understand why influence-guided generative correction improves downstream performance, we analyze the ranking induced by self-influence scores. While Fig.~\ref{fig:threshold} shows that corrupted and correctly labeled samples occupy distinct influence regions, Fig.~\ref{fig:ranking} evaluates how effectively this ranking prioritizes optimization-disruptive samples.
Fig.~\ref{fig:ranking} shows that corrupted samples are highly concentrated among the highest self-influence instances. Selecting only the top 5\% of the ranking recovers approximately one quarter of all corrupted samples, while the top 10\% and 15\% recover roughly one half and three quarters, respectively. Nearly all corrupted samples are recovered within the top 20\%, after which the recovery curve rapidly saturates, indicating that most optimization-disruptive supervision is concentrated within a relatively small subset of the training data.
Correctly labeled samples exhibit the opposite trend. Very few appear among the highest self-influence instances before the 20\% operating point, indicating that elevated self-influence primarily reflects supervision inconsistencies rather than intrinsically difficult examples. Beyond this point, increasingly many clean samples are included while only a few additional corrupted samples are recovered. Consequently, extending correction beyond the highest-ranked samples offers diminishing returns, increasing computational cost while unnecessarily modifying correctly supervised data.

These observations explain the behavior of DiffInf. Although 30\% synthetic label noise is introduced during training, only a subset of corrupted samples substantially perturbs optimization. Rather than attempting to correct every mislabeled instance, DiffInf focuses generative refinement on the highly influential samples that dominate decision-boundary formation. This targeted strategy preserves most correctly supervised data while maximizing the impact of correction, explaining its consistent improvements over both influence-based sample removal and conventional noisy-label learning methods in Table~\ref{tab:ffhq_classification}.

The proposed self-influence-guided generative correction substantially improves performance across all metrics, achieving 83.37\% accuracy, 94.94 AUROC, and $\kappa = 0.78$ for age, and 94.24\% accuracy, 99.38 AUROC, and $\kappa = 0.90$ for expression. Relative to noisy-label training, this corresponds to gains of 12.93 percentage points in accuracy, 9.46 points in AUROC, and 0.20 in $\kappa$ for age, and gains of 15.29 percentage points in accuracy, 4.95 points in AUROC, and 0.21 in $\kappa$ for expression. The consistent improvements across discrimination- and agreement-based metrics indicate that DiffInf enhances both class separability and label-consistent prediction reliability.

The comparison between self-influence removal and DiffInf is particularly informative because both methods rely on the same influence signal to identify problematic samples. For age prediction, removal improves performance to 81.23\% accuracy, 93.73 AUROC, and $\kappa = 0.76$, whereas DiffInf further improves these values to 83.37\%, 94.94, and 0.78. For expression recognition, removal achieves 93.42\% accuracy, 99.27 AUROC, and $\kappa = 0.88$, while DiffInf further improves performance to 94.24\%, 99.38, and 0.90. Although the margin over removal is smaller than the improvement over noisy training, it is consistent across both tasks and all reported metrics.

This behavior supports the central hypothesis of this work. Removing high-influence samples suppresses contradictory supervision but also reduces the effective sample size and can distort the empirical training distribution, particularly when influential samples correspond to rare visual regimes. In contrast, DiffInf preserves dataset cardinality while reducing the disruptive effect of these samples through targeted generative correction. Its consistent improvement over removal suggests that influential samples often contain valuable visual information that is better repaired than discarded.

DiffInf also compares favorably with established noisy-label baselines. For age classification, it outperforms Small\_loss, ELR+, proself, and promix across accuracy, AUROC, and $\kappa$. For expression recognition, DiffInf again achieves the best overall performance, exceeding the strongest competing baseline, self-influence removal, by 0.82 percentage points in accuracy, 0.11 AUROC, and 0.02 in $\kappa$. Perceptual similarity analysis (Table~\ref{tab:ffhq_lpips}) shows low LPIPS distances across age (0.219--0.227) and expression (0.196--0.244), indicating that correction remains within a constrained perceptual neighborhood of the original image. Thus, the observed improvements arise from targeted, attribute-consistent edits rather than aggressive image regeneration. Qualitative results in Figures~\ref{fig1} and~\ref{fig2} further support this observation: high self-influence samples exhibiting label--image mismatches are transformed into semantically aligned counterparts through localized modifications to attribute-relevant regions, such as skin texture for age and periocular or mouth configurations for expression. The absence of visible artifacts or identity drift further indicates that the diffusion latent space enables controlled semantic correction.

Taken together, the quantitative improvements, perceptual fidelity measurements, and qualitative evidence demonstrate that influence-guided generative repair produces a more coherent training distribution and stronger generalization than approaches based solely on loss-level regularization or sample filtering.

\section{Ablation Studies}
Table~\ref{tab:noise_results} evaluates the robustness of DiffInf across increasing levels of synthetic label corruption. At low-to-moderate noise rates (10--30\%), influence-guided generative correction (G) consistently outperforms simple removal of high-influence samples (R) across all metrics. The gains increase with corruption severity, improving accuracy from 80.08\% to 81.99\% and 83.37\% at 10\%, 20\%, and 30\% noise, respectively, while also achieving higher AUROC and Cohen's $\kappa$. The largest improvement occurs at 30\% noise, where generative correction exceeds removal by 2.14 percentage points in accuracy and 1.21 points in AUROC. These results support the central hypothesis of DiffInf: high-influence samples often contain valuable semantic information that is better repaired than discarded.

This trend reverses at the highest corruption level (40\%), where removal slightly outperforms generative correction. We hypothesize that correcting a large fraction of the training set introduces a distribution shift between generated and real images. Although the synthesized samples remain visually plausible, subtle generation artifacts or domain discrepancies may accumulate as their proportion increases, partially offsetting the benefits of recovering corrupted examples. In contrast, removal acts as a stronger regularizer under extreme corruption by excluding highly suspicious samples altogether. Nevertheless, the performance gap remains modest, indicating that DiffInf is reasonably robust even under corruption levels well beyond those typically encountered in curated facial attribute datasets. Overall, these findings suggest that influence-guided generative correction is most effective in the practically relevant low-to-moderate noise regime, where repairing suspicious samples consistently outperforms removal.

\begin{table}[t]
\centering
\caption{Performance under different noise levels.}
\label{tab:noise_results}
\scriptsize
\setlength{\tabcolsep}{3pt}

\resizebox{0.7\columnwidth}{!}{%
\begin{tabular}{ccccc}
\toprule
\textbf{Noise (\%)} & \textbf{Method} & \textbf{Acc.} & \textbf{AUROC} & $\boldsymbol{\kappa}$ \\
\midrule

\multirow{2}{*}{10}
& R & 79.246 & 92.504 & 0.71 \\
& G & \textbf{80.080} & \textbf{92.609} & \textbf{0.73} \\
\cmidrule(lr){1-5}

\multirow{2}{*}{20}
& R & 80.377 & 93.234 & 0.74 \\
& G & \textbf{81.992} & \textbf{93.930} & \textbf{0.75} \\
\cmidrule(lr){1-5}

\multirow{2}{*}{30}
& R & 81.230 & 93.730 & 0.76 \\
& G & \textbf{83.370} & \textbf{94.940} & \textbf{0.78} \\
\cmidrule(lr){1-5}

\multirow{2}{*}{40}
& R & \textbf{77.140} & \textbf{92.290} & \textbf{0.70} \\
& G & 75.440 & 91.650 & 0.68 \\
\bottomrule
\end{tabular}
}

\end{table}

\section{Conclusion and Future Work}
In this work, we introduced an influence-guided generative repair framework for learning under noisy supervision and demonstrated its effectiveness on multi-class facial attribute recognition. Rather than suppressing or discarding high self-influence samples, our approach uses diffusion-based latent editing to transform semantically misaligned instances into label-consistent counterparts. By restoring label--image alignment at the data level, the proposed framework produces a more coherent training distribution and improves generalization. Across both age and expression classification, DiffInf consistently improves accuracy, AUROC, and linearly weighted $\kappa$, indicating enhanced class separability, stronger ordinal consistency, and fewer severe cross-class errors while preserving perceptual fidelity, identity, and structural integrity.
A key conceptual contribution of this work is the reinterpretation of high-influence samples. Rather than treating high-influence samples as harmful noise, we show that they often contain valuable visual information despite semantically inconsistent supervision, and that repairing them instead of removing them preserves dataset diversity while reducing decision-boundary distortion, particularly in multi-class settings where mislabeled outliers can disproportionately perturb class transitions, thereby providing a mechanism for stabilizing learning under label corruption.

Several limitations warrant future investigation. First, the generative correction objective contains multiple interacting hyperparameters governing identity preservation, semantic alignment, and influence suppression, and broader ablations across tuning strategies and noise levels would better characterize its operating regime. Second, our influence signal relies on a first-order approximation and a surrogate predictor of high-influence membership rather than exact second-order influence; more accurate estimators could better distinguish mislabeled samples from genuinely difficult but correctly labeled instances. Third, demographic and subgroup-specific effects of generative correction require further study to ensure that repairing supervision inconsistencies does not inadvertently amplify bias. More broadly, the proposed paradigm extends naturally to medical imaging, fine-grained recognition, and other weakly supervised settings where labels are noisy but image content remains informative. By reframing noisy samples as candidates for semantic repair rather than removal, this work establishes a foundation for data-centric robustness strategies that unify attribution and generative modeling.

\bibliographystyle{splncs04}
\bibliography{main}
\end{document}